# Active Search for Sparse Signals with Region Sensing

**Yifei Ma**
Carnegie Mellon University
Pittsburgh PA 15213, US
yifeim@cs.cmu.edu

**Roman Garnett**
Washington University in St. Louis
St. Louis, MO, USA
garnett@wustl.edu

**Jeff Schneider**
Carnegie Mellon University
Pittsburgh PA 15213, US
schneide@cs.cmu.edu

## Abstract

Autonomous systems can be used to search for sparse signals in a large space; e.g., aerial robots can be deployed to localize threats, detect gas leaks, or respond to distress calls. Intuitively, search algorithms may increase efficiency by collecting aggregate measurements summarizing large contiguous regions. However, most existing search methods either ignore the possibility of such region observations (e.g., Bayesian optimization and multi-armed bandits) or make strong assumptions about the sensing mechanism that allow each measurement to arbitrarily encode all signals in the entire environment (e.g., compressive sensing). We propose an algorithm that actively collects data to search for sparse signals using only noisy measurements of the average values on rectangular regions (including single points), based on the greedy maximization of information gain. We analyze our algorithm in 1d and show that it requires $\tilde{O}(n/\mu^2 + k^2)$ measurements to recover all of $k$ signal locations with small Bayes error, where $\mu$ and $n$ are the signal strength and the size of the search space, respectively. We also show that active designs can be fundamentally more efficient than passive designs with region sensing, contrasting with the results of Arias-Castro, Candes, and Davenport (2013). We demonstrate the empirical performance of our algorithm on a search problem using satellite image data and in high dimensions.

## 1 Introduction

*Active search* describes the problem where an agent is given a target to search for in an unknown environment and actively makes data-collection decisions so as to locate the target as quickly as possible. Examples of this setting include using aerial robots to detect gas leaks, radiation sources, and human survivors of disasters. The statistical principles for efficient designs of measurements date back to Gergonne (1815), but the growing trend to apply automated search systems in a variety of environments and with a variety of constraints has drawn much research attention recently, due to the need to address the disparate aspects of new applications.

One possibility in such active search scenarios we aim to explore, inspired by the robotic aerial search setting but with statistical insights that we hope to generalize, is the opportunity to take aggregate measurements that summarize large contiguous regions of space. For example, an aerial robot carrying a radiation sensor will sense a region of space whose area depends on its altitude. How can such a robot dynamically trade off the ability to make noisier observations of larger regions of space against making higher-fidelity measurements of smaller regions?

To simplify the discussion, we will limit such *region sensing* observations to reveal the average value of an underlying function on a rectangular region of space, corrupted by independent observation noise. Noisy binary search is a simple realization of active search using such an observation scheme. This mechanism turns out to be sufficiently informative in the cases that we analyze to offer insights into a variety of search problems.

The ability to make aggregate region measurements in noisy environments has rarely been considered in previous work. *Bayesian optimization,* which has been used for localization of sparse signals (Carpin et al. 2015; Ma et al. 2015; Hernández-Lobato, Hoffman, and Ghahramani 2014; Jones, Schonlau, and Welch 1998), usually considers only point measurements of an objective function. Notice that point observations can be considered in our framework if the allowed region sensing actions are constrained to be arbitrarily small. On the other extreme, *compressive sensing* (Donoho 2006; Candès and Wakin 2008; Wainwright 2009), considers scenarios where every measurement can reveal information about the entire environment through linear projection with arbitrary coefficients. This is not always a realistic assumption, as for example for an aerial robot, which can only sense its immediate vicinity. Between the two extremes, Jedynak, Frazier, and Sznitman (2012); Rajan et al. (2015); Haupt et al. (2009); Carpentier and Munos (2012); Abbasi-Yadkori (2012); Yue and Guestrin (2011) considered policies for search where observations can be made on any arbitrary subset of the search space, including discontiguous subsets, which is also often incompatible with the constraints in physical search systems.

Another assumption we make, common for example in compressive sensing, is *sparsity.* We assume that there are only a small number of strong signals in the environment; our goal is to recover these signals. Sparsity is necessary for the definition of active search problems; otherwise, for dense or weak signals, there is usually no better search approach than simply exhaustively mapping the entire space.

In addition to applicability in real search settings, spar-sity has unique mathematical properties when considered alongside region sensing. In unconstrained sensing, Arias-Castro, Candes, and Davenport (2013) discovered a paradox that active compressive sensing (that is, the ability to adaptively select observations based on previously collected data) does not improve detection efficiency beyond logarithmic terms over random compressive sensing. This limitation is seen also when considering theoretical detection rates for active compressive sensing methods (Abbasi-Yadkori 2012; Carpentier and Munos 2012; Haupt et al. 2009). However, we show that active learning can in fact offer significant improvements in detection rates when observations are constrained to contiguous regions.

We propose an algorithm we call *Region Sensing Index* (RSI) that actively collects data to search for sparse signals using only noisy region sensing measurements. RSI is based on greedy maximization of information gain. Although information gain is a classic principle, we believe that its use in the recovery of sparse signals is novel and a good fit for robotic applications. We show that RSI uses $\tilde{O}(n/\mu^2 + k^2)$ measurements to recover all of $k$ true signal locations with small Bayes error, where $\mu$ and $n$ are the signal strength and the size of the search space, respectively (Theorem 3). The number of measurements with RSI is comparable with the rates offered by unconstrained compressive sensing, even though our constraints seem strong (i.e., region sensing loses all spatial resolution inside the region of measurement). Furthermore, we show that all passive designs under our contiguous region sensing constraint in $1d$ search spaces are fundamentally worse, with efficiency no better than sequential scanning of every point location, however strong the signals are. These results provide evidence to promote the use of and research into active methods.

## 1.1 Related Work

Arias-Castro, Candes, and Davenport (2013) proved that the minimax sample complexity[1] for any (i.e., potentially adaptive) algorithm to recover $k$ sparse signal locations is at least $\Omega(n/\mu^2)$, analyzing the problem in terms of the mean-squared error in the recovery of the underlying signal values. The authors also showed that a passive *random* design, combined with a nontrivial inference algorithm, e.g., Lasso (Wainwright 2009) or the Dantzig selector (Candes and Tao 2007), can have similar recovery rates (up to $O(\log n)$ terms). This result was presented as a paradox, suggesting that the folk statement that active methods have better sample complexity is not always true. Here we show that active search can make a substantial difference in recovery rates when the measurements are subject to the physically plausible constraint of region sensing, especially if the physical space has low dimensions.

Malloy and Nowak (2014) presented the first *active* search algorithm that achieves the minimax sample complexity for general $k$. The algorithm is called Compressive Adaptive Sense and Search (CASS) and it can be adapted to region sensing in one-dimensional physical spaces. CASS directly extends bisection search, by allocating different sensing budgets to measurements at different bisection levels so as to minimize the cumulative error rates. However, CASS may fail if the repeated measurements of the same regions do not contain perfectly independent noise. It also has the limitation that it requires knowledge of the sensing budget *a-priori*, yet produces no signal localization results until the very last measurements at the lowest level. Our paper addresses these practical issues with a redesigned active search algorithm using the Bayesian approach, which compares evidence instead of blindly trust the assumptions, and we use Shannon-information criteria, which implies bisection search in noiseless one-sparse cases.

---

[1]Sample complexity is equivalent to the number of measurements.

Braun, Pokutta, and Xie (2015) also used Shannon-information criteria for active search but did not analyze their sample complexity under noisy measurements. Jedynak, Frazier, and Sznitman (2012); Rajan et al. (2015) studied a similar search problem where the "regions" are relaxed to any unions of disjoint subsets.

## 2 Problem Formulation

Consider a discrete space that is the Cartesian product of one-dimensional grids, $\mathcal{X} = \prod_{i=1}^{d}[n_i]$; $[n] = \{1, \ldots, n\}$. Let $n = \prod n_i$ be the total number of points in $\mathcal{X}$ (here the product symbol is the arithmetic rather than the Cartesian product). We presume there is a latent real-valued nonnegative vector $\boldsymbol{\beta} \in \mathbb{R}^n$ that represents the vector of true signals at all locations in $\mathcal{X}$. We further assume that $\boldsymbol{\beta}$ is sparse: it has value $\mu > 0$ on $k \ll n$ locations in $\mathcal{X}$ and has value $0$ elsewhere. We consider making observations related to $\boldsymbol{\beta}$ through rectangular region sensing measurements, defined by

$$y_t = \mathbf{x}_t^\top \boldsymbol{\beta} + \varepsilon_t, \text{ s.t. } x_{tj} = w_t 1_{j \in A_t}, \ \varepsilon_t \sim \mathcal{N}(0, \sigma_t^2). \quad (1)$$

Here $\mathbf{x}_t \in \mathbb{R}^n$ is a sensing vector that has support on $A_t \subseteq \mathcal{X}$, a rectangular subset of $\mathcal{X}$. We assume that the sensing vector has equal weight $w_t$ across its support. The resulting measurement, $y_t$, is equal to the mean value of $\boldsymbol{\beta}$ on $A_t$ corrupted by independent Gaussian noise with variance $\sigma_t^2$. Note that selecting $A_t$ suffices to specify the measurement location.

In $1d$ search environments, $A_t$ may be any interval of $[n]$, and the corresponding design takes the form $\mathbf{x}_t = (0, \ldots, 0, w_t, \ldots, w_t, 0, \ldots, 0)^\top$. In higher search dimensions, we consider only regions that are contained in a hierarchical spacial pyramid (Lazebnik, Schmid, and Ponce 2006), i.e., a sequence of increasingly finer grid boxes with dyadic side lengths to cover the space at multiple resolutions.

Our goal is to choose a sequence of designs $\mathbf{X} = \{\mathbf{x}_t\}_{t=1}^{T}$ so as to discover the support of $\boldsymbol{\beta}$ with high confidence. Given a particular confidence, we will measure sample complexity by assuming $\|\mathbf{x}_t\|_2 = 1$ and $\sigma_t \equiv 1$ for each measurement and count the total number of measurements required to achieve that confidence, $T$. Letting $\|\mathbf{x}_t\|_2 = 1$ implies $w_t = 1/\sqrt{\|\mathbf{x}_t\|_0}$, which can be seen as a *relaxed* notion of the region average, because the signal strength of a region measurement, which is $\mu w_t$, still decreases as the region size $\|\mathbf{x}_t\|_0$ increases.

**Algorithm 1** Region Sensing Index (RSI)

**Require:** $\pi_0(k, n, \mu), T$ or $\epsilon$, and the unknown $\boldsymbol{\beta}^*$
**Ensure:** $\hat{S}_t$   // (5)
1: **for** $t = 1, 2, \ldots$ **do**
2:   pick $\mathbf{x}_t = \arg\max_{\mathbf{x} \in \mathcal{X}} I(\boldsymbol{\beta}; y \mid \mathbf{x}, \pi_{t-1})$   // (3)&(4)
3:   observe $y_t = \mathbf{x}_t^\top \boldsymbol{\beta}^* + \varepsilon_t$
4:   update $\pi_t(\boldsymbol{\beta}) \propto \pi_{t-1}(\boldsymbol{\beta}) p(y_t \mid \boldsymbol{\beta}, \mathbf{x}_{t-1})$   // (2)
5:   find $(\bar{\epsilon}_t, \hat{S}_t) = \arg\min_{|\hat{S}|=k} \frac{1}{k} \mathbb{E}[|\hat{S} \Delta S| \mid \pi_t]$   // (5)
6:   break if $t \geq T$ or $\bar{\epsilon}_t < \epsilon$, if either is defined

The measure of $T$ is made to be comparable with another common choice of sample complexity, the Frobenius norm of the entire design $\|\mathbf{X}\|_F^2$, when the rows of $\mathbf{X}$ are normalized (Arias-Castro, Candes, and Davenport 2013). However, the normalization is often overlooked in classical compressive sensing, which allows algorithms to cheat in region sensing by making an enormous number of measurements of small weight and changing the sensing locations frequently. Another measure of complexity is to measure both $\|\mathbf{X}\|_F^2$ and the number of location changes simultaneously (Malloy and Nowak 2014). However, our discretized counting of measurements is conceptually simpler.

Our analysis is Bayesian and we will analyze performance in expectation, with prior $\boldsymbol{\beta} \sim \pi_0(\boldsymbol{\beta})$, a uniform distribution on the model class, $\mathcal{S}_\mu\binom{n}{k}$, which includes all $k$-sparse models with $\mu$ signal strength among $n$ locations (i.e., it has $\binom{n}{k}$ possible outcomes). The Bayes risk will be measured by the expected Delta loss, $\bar{\epsilon}_T = \frac{1}{k}\mathbb{E}|S\Delta\hat{S}_T|$, where $\hat{S}_T$ is the best estimator of the $k$ signal locations after $T$ measurements and $\Delta$ is the symmetric difference operator on a pair of sets.

## 3 Proposed Methods

We note that region sensing loses all spatial resolution inside the region of measurement. Here we borrow ideas from noisy binary search, which has a similar property, and use information gain (IG) to drive the observation process. We name our algorithm *Region Sensing Index* (RSI, Algorithm 1). Like other active learning algorithms, RSI is a combination of an *inference* subroutine that constantly updates the distribution of $\boldsymbol{\beta}$ using the collected data and a *design* subroutine that chooses the next region to sense based on the latest information from the inference subroutine.

**The inference subroutine.** We use exact Bayesian inference with a uniform prior $\pi_0(\boldsymbol{\beta})$ on the model class $\mathcal{S}_\mu\binom{n}{k}$. Denote the outcome of the first $t$ measurements as $\mathcal{D}_t = \{(\mathbf{x}_\tau, y_\tau) : 1 \leq \tau \leq t\}$. Even though $\mathcal{D}_t$ contains a dependent sequence of data collections, where $\mathbf{x}_\tau$ depends on $\mathcal{D}_{\tau-1}, \forall \tau$, Bayesian inference decomposes into a series of efficient updates:

$$\pi(\boldsymbol{\beta} \mid \mathcal{D}_t) \propto \pi(\boldsymbol{\beta})p(\mathcal{D}_t \mid \boldsymbol{\beta})$$
$$= \pi_0(\boldsymbol{\beta}) \prod_{\tau=1}^t \left(p(\mathbf{x}_\tau \mid \mathcal{D}_{\tau-1})p(y_\tau \mid \boldsymbol{\beta}, \mathbf{x}_\tau)\right)$$
$$\propto \pi_0(\boldsymbol{\beta}) \prod_{\tau=1}^t p(y_\tau \mid \boldsymbol{\beta}, \mathbf{x}_\tau), \quad (2)$$

where $p(\mathbf{x}_\tau \mid \mathcal{D}_{\tau-1})$ is the design without knowledge of the true $\boldsymbol{\beta}$ and thus dropped. Define $\pi_t(\boldsymbol{\beta}) = \pi(\boldsymbol{\beta} \mid \mathcal{D}_t)$; the updates have the form $\pi_t(\boldsymbol{\beta}) \propto \pi_{t-1}(\boldsymbol{\beta})p(y_t \mid \boldsymbol{\beta}, \mathbf{x}_t) = \pi_{t-1}(\boldsymbol{\beta})\phi(y_t - \mathbf{x}_t^\top \boldsymbol{\beta})$, where $\phi$ is the standard normal pdf.

**The design subroutine.** The next sensing vector, $\mathbf{x}_{t+1} \in \mathcal{X}$, is chosen to maximize the IG:

$$I(\boldsymbol{\beta}; y \mid \mathbf{x}, \pi_t) = H(y \mid \mathbf{x}, \pi_t) - \mathbb{E}\big[H(y \mid \mathbf{x}, \boldsymbol{\beta}) \mid \pi_t\big], \quad (3)$$

which is the difference between the entropy of the marginal distribution, $p(y \mid \mathbf{x}, \pi_t) = \int \phi(y - \mathbf{x}^\top \boldsymbol{\beta})\pi_t(\boldsymbol{\beta})\,\mathrm{d}\boldsymbol{\beta}$, and the expected entropy of the conditional distribution, $p(y \mid \boldsymbol{\beta}; \mathbf{x}) = \phi(y - \mathbf{x}^\top \boldsymbol{\beta})$. The latter, i.e., the conditional distribution for any realization of $\boldsymbol{\beta}$, has fixed entropy: $\log\sqrt{2\pi e}$. Meanwhile, the marginal entropy has no closed-form solutions; instead, we use numerical integration.

The numerical integration is rather straightforward, because the marginal *density* function is analytical. From now on, we will assume that $(\mathbf{x}, A, a, w_\mathbf{x})$ correspond to the same design (its sensing vector, its locations, its region size, and its sensing weight per coordinate, respectively). Define two new variables, $\lambda = \mu w_\mathbf{x} (= \mu/\sqrt{a})$ and $\gamma = \mathbf{x}^\top\boldsymbol{\beta}/\lambda$, and one new parameter $\mathbf{p} = (p_0, \ldots, p_k)^\top$ in (4). The goal is to change the variable of the integration for the marginal *density* function of $y$ to:

$$p(y \mid \mathbf{x}, \pi_t) = \int \pi_t(\boldsymbol{\beta})\phi(y - \mathbf{x}^\top\boldsymbol{\beta})\,\mathrm{d}\boldsymbol{\beta}$$
$$= \sum_{c=0}^k p_c\,\phi(y - c\lambda) = p(y \mid \lambda, \mathbf{p}),$$
where $\quad p_c = \Pr(\gamma = c) = \sum_{\boldsymbol{\beta}:\mathbf{x}^\top\boldsymbol{\beta}=c\lambda} \pi_t(\boldsymbol{\beta}). \quad (4)$

Notice, $\gamma$ only has a finite number of choices: $\gamma = |A \cap S| \in \{0, \ldots, k\}$, where $S$ is the nonzero support of $\boldsymbol{\beta}$, because both $\mathbf{x}$ and $\boldsymbol{\beta}$ are constant on their respective supports ($x_j = w_\mathbf{x}, \forall j \in A$ and $\beta_j = \mu, \forall j \in S$). We then numerically evaluate $H(y \mid \mathbf{x}, \pi_t) = H(y \mid \lambda, \mathbf{p})$ with the obtained (4).

**The Bayes estimator of signal locations.** We pick the $k$-sparse set $\hat{S}_T$ to minimize the posterior risk:

$$\min_{|\hat{S}|=k} \frac{1}{k}\mathbb{E}\big[|\hat{S}\Delta S| \mid \pi_T\big] = \frac{1}{k}\sum_{\hat{i}\in\hat{S}} \mathbb{E}\big(\mathbf{1}_{\{\beta_{\hat{i}}=0\}} \mid \pi_T\big), \quad (5)$$

where $\beta_{\hat{i}}$ is the $\hat{i}$-th element of $\boldsymbol{\beta}$. In other words, RSI picks the top $k$ locations where the posterior marginal expectation is the largest. When $k = 1$, this is equivalent to picking $\hat{\boldsymbol{\beta}}_T = \arg\max \pi_T(\boldsymbol{\beta})$. Otherwise, (5) yields the smallest Bayes risk $\bar{\epsilon}(\mathcal{D}_T)$ given any collected data $\mathcal{D}_T$.

### 3.1 Accelerations

In practice, holding $\binom{n}{k}$ models in memory can be infeasible if $k$ is large, we can instead recover the support of $\boldsymbol{\beta}$ element-wise by repeatedly applying RSI assuming $k = 1$. After the posterior distribution $\pi_t(\boldsymbol{\beta}^{(1)})$ converges to a point-mass distribution at the most-likely one-sparse model with sufficient confidence, we report its location and move on by

---
[2] In real world experiments, we additionally estimate $\hat{\mu}_{\hat{j}}$ using a point measurement on the inferred signal location for better modeling.

Table 1: Conditions and conclusions for sample complexity.

| Design Type | Region Sensing | Algorithm | Prior for Bayes Risk | Min $T$ to Guarantee $\bar{\epsilon}_T = \frac{1}{k}\mathbb{E}|S\Delta\hat{S}_T| \leq \epsilon$ | Sample Complexity* |
|---|---|---|---|---|---|
| passive | yes | (any) | $\pi_0$ ($\mu \to \infty$) | $T \geq \frac{n}{2}(1 - \frac{n-1}{n-k}\epsilon)$ (Theorem 1) | $\Theta(n)$ |
|  |  | Point sensing |  | $T \leq n(1 - \frac{n-1}{n-k}\epsilon)$ (Corollary A.2) |  |
| active | no | (any) | $\tilde{\pi}_0$ | $T \geq \frac{4n}{\mu^2}(1 - \epsilon)^2$ (Theorem 2) | $\Omega(\frac{n}{\mu^2})^\dagger$ |
|  | yes | CASS [2014] | max risk (incl. $\pi_0$) | $T \leq 20\frac{n}{\mu^2}\log(\frac{8k}{\epsilon}) + 2k\log_2(\frac{n}{k})$ | $\tilde{O}(\frac{n}{\mu^2} + k)^\ddagger$ |
|  |  | RSI (ours) | $\pi_0$ | $\bar{T}_\epsilon \leq 50(\frac{n}{\mu^2} + \frac{k^2}{9})\log_2(\frac{2}{\epsilon})\log(\frac{n}{\epsilon})$ (Theorem 3) | $\tilde{O}(\frac{n}{\mu^2} + k^2)^\ddagger$ |

* Assume $\epsilon = O(1)$ and $k \ll n$. † Shown for unconstrained sensing; binary search requires $\Omega(\log_2(n) + k)$ additional measurements. ‡ $\log(n)$ terms are left out. $\bar{T}_\epsilon$ is defined differently; see Section 4.2 for details.

---

**Algorithm 2** Region Sensing Index-Any-$k$ (RSI-A)

**Require:** $n, \mu, \epsilon$, and the unknown $\boldsymbol{\beta}^*$
**Ensure:** $\hat{S}$
1: initialize $\hat{S} = \emptyset$, $\hat{\boldsymbol{\beta}} = \mathbf{0}$
2: **for** $k = 1, 2, \ldots$, **do**
3:   infer $\pi_0(\boldsymbol{\beta}^{(k)}) \propto \prod_{\tau=1}^{t} p(y_\tau \mid \boldsymbol{\beta}^{(k)} + \hat{\boldsymbol{\beta}}, \mathbf{x}_\tau)$,
    $\forall \boldsymbol{\beta}^{(k)} \in \{\mu\mathbf{1}_j : j \notin \hat{S}\}$
4:   call $\hat{S}^{(k)} = $ RSI $(\pi_0, \epsilon, \boldsymbol{\beta}^* - \hat{\boldsymbol{\beta}})$
5:   aggregate $\hat{S} = \cup_{c \leq k}\hat{S}^{(c)}$ and $\hat{\boldsymbol{\beta}} = \sum_{\hat{j} \in \hat{S}} \hat{\mu}_{\hat{j}} \mathbf{1}_{\hat{j}}$.[2]

---

removing the reported point from the search and recomputing the posterior distributions using the uniform prior, $\pi_0(\boldsymbol{\beta}^{(2)})$, on the new class, $\mathcal{S}_\mu\binom{n-1}{1}$.

We call this alternative algorithm *Region Sensing Index-Any-$k$* (RSI-A, Algorithm 2) and use it in our simulations so that the computational cost is no longer exponential in $k$. Notice, our analysis is for the unmodified RSI; the statistical disadvantage of RSI-A is no more than $O(k)$, multiplicatively.

When implementing RSI-A, we also avoid unnecessary numerical integration (3), if the region is guaranteed to have inferior IG, indicated by its **p** vector (4), which is easier to compute. We use the fact that $I(\gamma; y \mid \mathbf{p}, \lambda)$ with fixed $\lambda > 0$ is concave in the probability simplex $\Delta^k = \{\mathbf{p} \in [0, 1]^{k+1} : \mathbf{p}^\top \mathbf{1} = 1\}$. Under $k = 1$ approximation, the region whose marginal probability $p_1 = \sum_{\mathbf{x}^\top \boldsymbol{\beta} > 0} \pi(\boldsymbol{\beta})$ is closest to 0.5 will provably have the largest IG among all regions of the same size. Thus, we find the region with the highest IG in two steps: (1) compare the $p_1$ value for all regions for every region size and (2) evaluate the IG of only these regions with the best $p_1$ values (closest to 0.5) in their region sizes.

## 4 Theoretical Analysis in 1D

The analysis is cleanest when the search space is 1d, where the regions can be any integer intervals that subset $[1, n]$. Without loss of generality (WLOG), assume $n$ is a multiple of $k$ and $n \geq 2k$. Our goal is to find the smallest number of measurements, $T$, to guarantee a small Bayes risk $\bar{\epsilon}_T = \frac{1}{k}\mathbb{E}|S\Delta\hat{S}_T| \leq \epsilon$. Table 1 summarizes our analysis. The sample complexity is best appreciated assuming $\mu \gg 1$, $k \ll n$, and $\epsilon = \mathcal{O}(1)$. A typical choice is $\epsilon = 1/2$, i.e., the number of measurements to guarantee that half of the signal support can be recovered on average.

### 4.1 Baseline Results

Here we provide lower bounds on sample complexity. We show that under region-sensing constraints, all passive methods require $T \geq \Omega(n)$ measurements and active methods require $T \geq \Omega(n/\mu^2 + k)$. When $\mu \gg 1$, active methods have significant potential for improvement over passive methods using region sensing, which contradicts with the view in unconstrained compressive sensing by Arias-Castro, Candes, and Davenport (2013); Soni and Haupt (2014).

**Theorem 1** (Limits of any passive methods using region sensing). *Assume $\boldsymbol{\beta}$ has prior $\pi_0$ (uniform random on $\mathcal{S}_\mu\binom{n}{k}$). Any passive method with $T$ noiseless region measurements on 1d must incur Bayes risk $\bar{\epsilon}_T \geq \frac{n-k}{n-1}(1 - \frac{2T}{n})$. To guarantee $\bar{\epsilon}_T \leq \epsilon$, $T \geq \frac{n}{2}(1 - \frac{n-1}{n-k}\epsilon)$ is required.*

The proof is due to model identifiability, neglecting observation noise. More details can be found in the appendix. It applies to any $\mu \geq 0$ and particularly $\mu \to \infty$.

**Theorem 2** (Limits of any methods, (Arias-Castro, Candes, and Davenport 2013)). *Assume $\boldsymbol{\beta}$ has a slightly different prior, $\tilde{\pi}_0$, that includes each location in $\mathcal{X}$ in the support of $\boldsymbol{\beta}$ independently with probability $k/n$. Any method (including active and non-region-sensing) must have $\bar{\epsilon}_T \geq 1 - \frac{\mu}{2}\sqrt{T/n}$. To guarantee $\bar{\epsilon}_T \leq \epsilon$, $T \geq \frac{4n}{\mu^2}(1 - \epsilon)^2$ is required.*

The proof can be found under Theorem 3 of (Arias-Castro, Candes, and Davenport 2013). Arias-Castro, Candes, and Davenport (2013) gave a minimax risk with similar terms by modifying $\tilde{\pi}_0$ to a *least favorable prior* on all models that are at most $k$-sparse. However, we only study Bayes risk for technical convenience.

When using Theorem 2 for reference, notice the difference between $\tilde{\pi}_0$ and $\pi_0$ that the former additionally treats the sparsity to be a random variable $\tilde{k}$ with expectation $k$. From concentration inequalities, $|\tilde{k} - k| \leq O(\sqrt{k})$, with high probability. While $\tilde{k}$ and $k$ are not directly comparable, Theorem 2 is still a useful baseline. Under region-sensing constraints, the number of measurements must also be at least $\Omega(k)$ to allow visits to most of the nonzero locations at least once, in a nontrivial draw of $S$ where the signals are separated.

With respect to Theorem 1, the point sensing or any non-repeating region sensing will achieve the optimal sample complexity (up to constant factors, see Appendix A for more details). For Theorem 2, the CASS method published by Malloy and Nowak (2014) for active sensing with region constraints[3] acheives a nearly optimal rate in theory. Table 1 contains a detailed summary of the sample complexities of several algorithms, including our own.

### 4.2 Main Result

For technical convenience, we directly express our main result in terms of the expected number of measurement that are actually taken so as to realize $\bar{\epsilon}(D_\mathcal{T}) \leq \epsilon$ for a given threshold $\epsilon$ in an experiment. Taking $\mathcal{T} = \mathcal{T}_\epsilon$ as a random variable, the expected number of actual measurements is different from the pre-determined sampling budget that an algorithm fully consumes to guarantee a desirable averaged risk (see Section 4.1). However, it is a comparable alternative in Bayesian analysis, used by e.g., Lai and Robbins (1985); Kaufmann, Korda, and Munos (2012). When the objective is constant $\epsilon = \mathcal{O}(1)$, our result implies a deterministic budget requirement of the same order of complexity, $T \leq \epsilon_2^{-1} \mathbb{E} \mathcal{T}_{\epsilon_2}$, where $\epsilon_2 = \frac{\epsilon}{2}$, by direct application of Markov's inequality.

**Theorem 3** (Sample complexity of RSI). *In active search for $k$ sparse signals with strength $\mu$ in 1d physical space of size $n \geq 2k$ (WLOG, assume $n$ is a multiple of $k$), given any $\epsilon > 0$ as tolerance of posterior Bayes risk,* RSI *using region sensing has bounded expected number of actual measurements,*

$$\bar{T}_\epsilon = \mathbb{E}[\min\{\mathcal{T} : \bar{\epsilon}(D_\mathcal{T}) \leq \epsilon\}]$$
$$\leq 50 \left(\frac{n}{\mu^2} + \frac{k^2}{9}\right) \log_2\left(\frac{2}{\epsilon}\right) \log\left(\frac{n}{\epsilon}\right) = \tilde{O}\left(\frac{n}{\mu^2} + k^2\right), \quad (6)$$

*where the expectation is taken over the prior distribution and sensing outcomes.*

### 4.3 Proof Sketch

The proof for Theorem 3 hinges on an observation that the information gain (IG) where RSI makes measurements is consistently large, before active search terminates with minimal Bayes risk. For example, the IG of any measurement in binary search with $k = 1$ and noiseless observations is always $O(\log(2))$. However, IG is harder to approximate when the observations are noisy. Therefore, we first show an intuitive lower bound for IG. Recall notations from (4).

---
[3]The original result in Malloy and Nowak (2014) is stronger; it considers the maximum probability of support recovery mistakes, $P(S \neq \hat{S}) \leq \delta$, for any $S$ that are $k$-sparse and any signals with at least $\mu$ strength.

**Proposition 4.** *The IG score of a region sensing design has lower bounds with respect to its design parameters $(\lambda, \mathbf{p})$, as*

$$I(\gamma; y \mid \lambda, \mathbf{p}) \geq 2q_c \bar{q}_c \left(2\Phi\left(\frac{\lambda}{2}\right) - 1\right)^2$$
$$\geq \frac{1}{12} \min\{q_c, \bar{q}_c\} \min\{\lambda^2, 3^2\}, \quad \forall 1 \leq c \leq k, \quad (7)$$

*where $q_c = \Pr(\gamma \geq c) = \sum_{\kappa \geq c} p_\kappa$, $\bar{q}_c = 1 - q_c$, and $\Phi(u)$ is the standard normal cdf.*

The proof uses Pinsker's inequality and is given in Section B in the appendix. Notice using the common choice of Jensen's inequality will give bounds in the opposite direction. To formalizes our observation that the IG is bounded:

**Lemma 5.** *WLOG, assume $n$ is a multiple of $k$ and $n \geq 2k$. At any step, if the current Bayes risk $\bar{\epsilon}(D) > \epsilon$, we can always find a region $A$ of size at most $\frac{n}{k}$, such that $\lambda^2 \geq \frac{\mu^2}{a} = \frac{k\mu^2}{n}$ and $\frac{\epsilon}{2} \leq \mathbb{E}[\gamma \mid D] \leq 1 - \frac{\epsilon}{2}$ (we call this* Condition E*), which further yields*

$$I(\gamma; y \mid \lambda, \mathbf{p}) \geq I_\epsilon^* = \frac{\epsilon}{25k} \min\left\{\frac{k^2 \mu^2}{n}, 3^2\right\}. \quad (8)$$

The way to find the region $A$ that satisfies *Condition E* is given in Lemma B.5 in the appendix. The reason that Condition E is sufficient for (8) can be derived from Proposition 4 for $k = 1$ and Lemma B.6 in the appendix for $k > 1$. □

Eq (8) shows the minimum decrease in the model entropy in expectation after each measurement, starting from the maximum entropy of a uniform prior distribution, $k \log(n)$. However, the posterior entropy can never be negative, which implies a bound on the expected number of times that (8) can be applied, i.e. the expected number of measurements to reach $\epsilon$ Bayes risk is $\frac{25 \log(n)}{\epsilon}\left(\frac{n}{\mu^2} + \frac{k^2}{9}\right)$. Lemma D.5 in the appendix shows some additional improvements to obtain the logarithmic dependency of $\epsilon$ in Theorem 3.

## 5 Simulation Studies

We evaluated RSI or its approximation RSI-A when $k > 1$. Other baseline algorithms include:

• **CASS** (compressive adaptive sense and search) (Malloy and Nowak 2014): a branch-and-bound algorithm that traverses the region hierarchy from top to down using pre-allocated budgets per level. We count each $\mathbf{x}_i$ as $\|\mathbf{x}_i\|_2^2$ region sensing measurements (rounded up to the next integer).

• **Point** sensing: a passive design that uses exhaustive point measurements on all locations.

• **CS** (compressive sensing) (Donoho 2006): a non-region-sensing design that draws $\mathbf{x}_t \sim \mathcal{N}(\mathbf{0}, \mathbf{I})$ and rescales $\|\mathbf{x}_t\|_2^2$ to 1. CS then solves a convex optimization problem to infer the nonzero signals, by minimizing $\sum_t \|y_t - \mathbf{x}_t^\top \boldsymbol{\beta}\|_2^2 + \lambda \|\boldsymbol{\beta}\|_1$ s.t. $\boldsymbol{\beta} \geq 0$, where $\lambda$ is chosen to produce exactly $k$ nonzero coefficients using the Lasso regularization path.

We picked $n = 1024$ and various $k$ (sparsity) and $d$ (the dimension of the physical space) annotated below the plots. In the $d = 5$ case, we chose the region space to be the Cartesian product of $[4]^5$ and allowed regions from a spatial pyramid (Lazebnik, Schmid, and Ponce 2006) of granularity

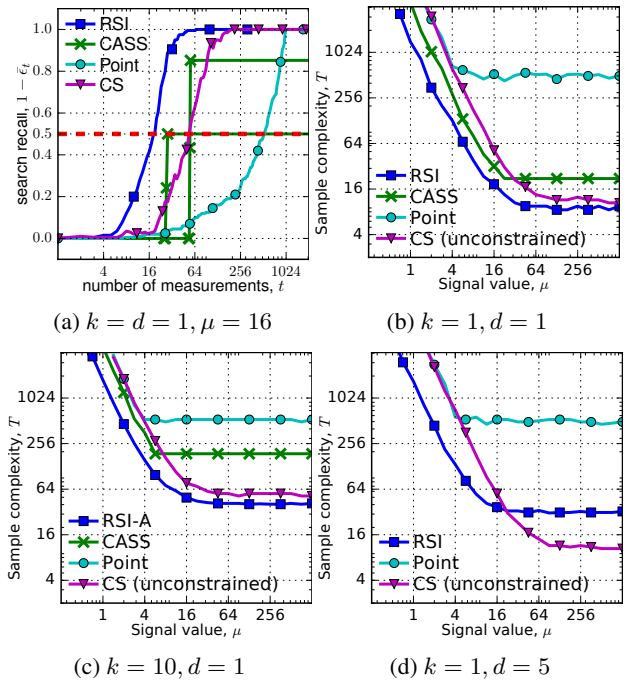

Figure 1: Sensing efficiency. (a) Average search progresses as more measurements are taken. (b-d) Minimum sample size $T$ in different SNR scenarios to guarantee $\bar{\epsilon}_T < 0.5$.

$4^5$, $2^5$, and $1^5$. Each method was run with 200 repetitions to find its average performance.

Figure 1(a) compares the recall rates of the algorithms as they progressed in a 1d search for a single true signal of strength $\mu = 16$. RSI was the most efficient, finding the correct location in $50\%$ of the cases with as few as $T = 20$ measurements. CASS was comparable only at the step points when all the allocated budgets were used, due to its rather rigid designs. We drew multiple curves for CASS to reflect this fact; the turning points were at $T = 28$ and $56$ for $\epsilon = 0.5$ and $0.85$, respectively. CS was less effective compared with CASS with equal budgets (e.g., $\|\mathbf{X}\|_F^2 = 52 > 28$ for $\epsilon = 0.5$) which agrees with the analysis in Arias-Castro, Candes, and Davenport (2013). Point sensing was the least efficient, using $T = n/2 = 512$ measurements, which was worse than the other methods by a factor of $\tilde{\Omega}(\mu^2)$ (ignoring logarithmic terms). Notice, due to non-identifiability, any passive designs would have equal or worse rates.

Figure 1(b) extends the comparison on the full spectrum of SNR, $1/4 < \mu < 1024$, showing the minimum number of measurements $T$ to guarantee constant Bayes risk $\bar{\epsilon}_T < 0.5$. RSI led the comparison, showing a sample complexity of $\tilde{O}(n/\mu^2)$ when $\mu$ is small and $\tilde{O}(1)$ when $\mu$ is large. CASS also had a similar trend. CS ignores the region sensing constraints and was inferior to RSI. Notice CS also has a minimum sample complexity, but in order to meet the incoherence conditions for Lasso sparsistency (Candes and Tao 2007; Wainwright 2009; Raskutti, Wainwright, and Yu 2010), the rank of the covariance matrix of the measurements $\mathbf{X}_S^\top \mathbf{X}_S$ must be at least $k$. Point sensing and other passive region sensing would always require at least $\Omega(n)$ measurements regardless of $\mu$. Figure 1(c-d) show similar conclusions with other choices of $k$ and $d$. The number of measurements was largely unaffected by $k > 1$ if $\mu$ is low, which supports the first term of Theorem 3, which is $\tilde{O}(n/\mu^2)$. Comparisons between CS and RSI in high dimensions ($d > 1$) depend on how region constraints are defined. In our high-dimensional simulations, the region choices were rather limiting for RSI, giving more advantage to the unconstrained CS when $\mu$ is large.

## 6 Real World Datasets

Region sensing was intended to address the problems of real robotic search. Here, we took satellite images like Figure 2a and used natural blue pixels, e.g., the blue roof which we circled near the lower left of the center of the image, as a simulated target of interest. These experiments directly simulate search and rescue in open areas based on life jacket colors or communication signals and also share similarities with gas leaks or radiation detection, where real data is usually sensitive or expensive. Many assumptions were violated in these experiments, e.g., the noise was not iid and the target was a collection of neighboring pixels. For the purpose of more accurate modeling of the actual measurement powers at each region level, we used statistics from the real data to model $\mu w(a)$ and $\sigma(a)$ as functions of the region size $a(= \|\mathbf{x}\|_0)$.

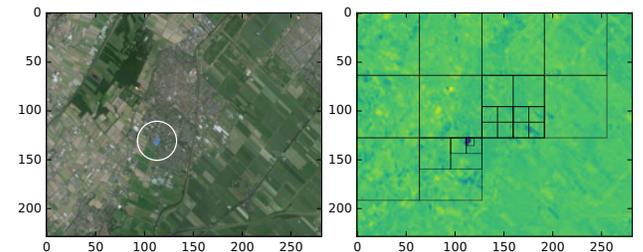

(a) Target blue pixels (circled)  (b) RSI search pattern

Figure 2: Demo active search on real images.

Figure 2b shows in the background the actual scalar observations, affinely transformed from the original RGB values to filter out the target blue color. The foreground contains the rectangular regions of measurement, sequentially decided by RSI after observing the average values in previous regions. Feasible region choices were contained in a spatial pyramid (Lazebnik, Schmid, and Ponce 2006). RSI behaved similarly to sequential scanning at the optimal altitude except for occasional bisections into subregions.

By comparing the IG of all feasible regions, RSI usually decides to (a) sense the next region in space when the previous outcome is low, (b) investigate the subregions when the last parent region yields a large outcome (we disallow repeated actions for the lack of noise-independence,) or (c) back out from an investigation if the subsequent measurements yield low outcomes. Option (c) demonstrates the ability of error recovery, which is our advantage to CASS thanks to Bayesian modeling. The search in Figure 2b ended after 36 measurements, whereas the image contained 36 000 pixel points.

Figure 3 compares the performances on 221 image patches of $512 \times 512$ pixels, cropped from National Agriculture Imagery Program (NAIP).[4] The other algorithms for comparison include random (point), CS, and CASS*. Here, CASS* is a modified CASS method where each measurement can only be taken once, because repeated measurements yield the same outcome. To fully represent CASS*, in addition to choosing $k$ by the true sparsity, we added fixed choices of $k = 64$ and $512$, yielding three different curves.

RSI achieved the best performance, finding on average $60\%$ blue pixels with as few as $1700$ measurements ($0.5\%$ of the total number of feasible observations). CASS* performance highly depended on the parameter choices and produced results only near the end of the experiment. CS did poorly, probably due to the fact the signals were not iid (a blue object can contain multiple pixels).

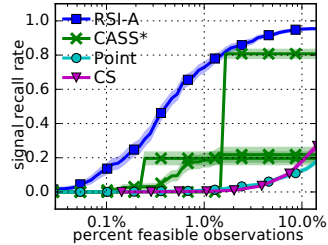

Figure 3: Performances on 221 NAIP image crops.

## 7 Discussions

Region sensing is a new setting motivated by robotic search operations where we also found statistical insights to contrast with the unconstrained sensing in Arias-Castro, Candes, and Davenport (2013). RSI performs near-optimally in 1d search domains and fundamentally faster than passive sensing. In higher dimensions, the analysis may be harder, especially for passive baselines. The number of subregions generated by intersecting the measurement regions may be harder to count, unless measurement regions are restricted to grid regions in a spatial pyramid (such that any pair of regions is either nested or disjoint). We also want to establish frequentist analysis in the future. Finally, it is interesting to generalize the measurement model beyond taking the average value of a single region at a time.

## Acknowledgments

This work is partially supported by the DARPA grant FA87501220324, National Science Foundation under Award Number IIA-1355406, and ARPA-E TERRA-REF award DE-AR0000594. We also appreciate suggestions and discussions from Aarti Singh, Ying Yang, and Yining Wang.

---
[4] https://lta.cr.usgs.gov/node/300

# Active Search for Sparse Signals with Region Sensing (Appendix)

**Yifei Ma**
Carnegie Mellon University
Pittsburgh PA 15213, US
yifeim@cs.cmu.edu

**Roman Garnett**
Washington University in St. Louis
St. Louis, MO, USA
garnett@wustl.edu

**Jeff Schneider**
Carnegie Mellon University
Pittsburgh PA 15213, US
schneide@cs.cmu.edu

## Abstract

This supplementary material includes both theoretical details (**Section A–B**) and additional empirical results (**Section C**). For the theoretical part, our main paper has two separate results: a hardness result showing passive methods under region sensing constraints in 1d search spaces cannot be efficient (Theorem 1 and the first row in Table 1 in the main paper) and a positive result showing that our *Region Sensing Index* (**RSI**) is sample-efficient (Theorem 3), comparable with the optimal efficiency obtained in (Arias-Castro, Candes, and Davenport 2013). The empirical part contains the choice of parameters and a demo of search results. For convenience, in the appendix, we sometimes use $K$ to represent the total number of sparse signals in the system, whereas in the main document we always used $k$.

## A  Theoretical Properties for Passive Sensing

**Theorem A.1** (**Theorem 1** in the main document; limits of any passive methods using region sensing). *Assume $\beta$ has prior $\pi_0$ (uniform random on $\mathcal{S}_\mu \binom{n}{k}$). Any passive method with $T$ noiseless region measurements on 1D must incur Bayes risk $\bar{\epsilon}_T \geq \frac{n-k}{n-1}(1 - \frac{2T}{n})$; to guarantee $\bar{\epsilon}_T \leq \epsilon$, it requires $T \geq \frac{n}{2}(1 - \frac{n-1}{n-k}\epsilon)$.*

*Proof.* We count the number of non-identifiability models with $T$ noiseless observations, particularly when $T < \frac{n}{2}$.

An aggregate measurement on region $[a_i, b_i) \subset [1, n+1)$ cannot identify the sparse support inside $[a_i, b_i)$ (or its complement), unless it intersects with another aggregate measurement. Should two measurement regions intersect, the model is still non-identifiable inside the intersection, set differences, and the complement of the union of both. To find out the set of all disjoint subsets where the model is non-identifiable given any passive design with $m$ region measurements, $\{[a_i, b_i) \subset [1, n+1) : i = 1, \ldots, m\}$, we simply sort the unique end points as $c_1 < \cdots < c_p \in \{a_i, \ldots, a_m\} \cup \{b_1, \ldots, b_m\}$, where $p \leq 2m$, and use the following set of $p$ elementary subsets:

$$\{\underbrace{[c_j, c_{j+1})}_{C_j} : j = 1, \ldots, p-1\} \cup \{\underbrace{[c_p, n+1) \cup [1, c_1)}_{C_p}\}, \tag{A.1}$$

where the last subset is created to ensure that the number of sparse supports in the full set equals $k$. Notice, (A.1) is also the largest set of disjoint subsets that can be created using intersections, unions, and complements on the regions of measurements.

We will continue our discussion assuming that the measurements are made on the subsets contained in (A.1). When the observations are noiseless, (A.1) is a superior design than the original design, whose outcomes can be inferred as

$$\mathbf{x}_{[a_i, b_i)}^\top \beta = \sum_{j=1}^{p-1} \frac{c_{j+1} - c_j}{b_i - a_i} \mathbf{x}_{[c_j, c_{j+1})}^\top \beta. \tag{A.2}$$

At this point, it is easy to see that the minimum sample size to guarantee that the signals can be fully identifiable in the worst case is $T \geq \frac{n}{2}$; the necessary (and sufficient) condition is to have $|C_j| = 1, \forall j = 1, \ldots, p$, which requires $2T \geq p \geq n$.

For $\epsilon > 0$, we compute the expected Delta-risk given any fixed design which yields $p$ elementary subsets as shown in (A.1). Let $n_j = |C_j|, j = 1, \ldots, p$. If the model $\beta$ distributes $k_j$ supports in subset $C_j$, respectively, then on any region where $n_j > k_j > 0$, the inference algorithm can only make a random guess, e.g., for the first $k_j$ elements. Let $\boldsymbol{\beta}_{C_j}$ be the signal vector on subset $C_j$, the conditional expected error on this subset is:

$$\mathbb{E}\big[|\boldsymbol{\beta}_{C_j} \Delta \hat{\boldsymbol{\beta}}_{C_j}| \mid k_j\big] = \sum_{e_j=1}^{k_j} \frac{\binom{n_j-k_j}{e_j}\binom{k_j}{k_j-e_j}}{\binom{n_j}{k_j}} e_j = \sum_{e_j=1}^{k_j} \frac{\binom{n_j-k_j-1}{e_j-1}\binom{k_j}{k_j-e_j}}{\binom{n_j}{k_j}} (n_j - k_j)$$

$$= \frac{\binom{n_j-1}{k_j-1}}{\binom{n_j}{k_j}}(n_j - k_j) = \frac{k_j(n_j - k_j)}{n_j}. \quad (A.3)$$

The total risk conditioned on all of $k_j : j = 1, \ldots, p$ is:

$$\mathbb{E}\big[|\boldsymbol{\beta}\Delta\hat{\boldsymbol{\beta}}| \mid k_1, \ldots, k_p\big] = \sum_{j=1}^{p} \mathbb{E}\big[|\boldsymbol{\beta}_{C_j}\Delta\hat{\boldsymbol{\beta}}_{C_j}| \mid k_j\big] = \sum_{j=1}^{p} \frac{k_j(n_j - k_j)}{n_j}. \quad (A.4)$$

Using the law of total expectation assuming $\boldsymbol{\beta}$ to be uniformly distributed, we can compute the expected error of the given passive design as

$$\mathbb{E}|\boldsymbol{\beta}\Delta\hat{\boldsymbol{\beta}}| = \sum_{k_1+\cdots+k_p=K} \left(\frac{\prod_{j=1}^{p}\binom{n_j}{k_j}}{\binom{n}{K}}\right)\left(\sum_{j=1}^{p}\frac{k_j(n_j-k_j)}{n_j}\right)$$

$$= \sum_{j=1}^{p} \sum_{k_1+\cdots+k_p=K} \frac{\prod_{j'=1}^{p}\binom{n_{j'}}{k_{j'}}}{\binom{n}{K}} \frac{k_j(n_j-k_j)}{n_j}$$

$$= \sum_{j=1}^{p} \sum_{k_1+\cdots+k_p=K} \frac{(n_j-1)\binom{n_j-2}{k_j-1}\prod_{j'\neq j}\binom{n_{j'}}{k_{j'}}}{\binom{n}{K}}$$

$$= \sum_{j=1}^{p} \frac{(n_j-1)\binom{n-2}{K-1}}{\binom{n}{K}} = \frac{(n-p)\binom{n-2}{K-1}}{\binom{n}{K}} = \frac{K(n-K)}{n(n-1)}(n-p) \leq K\frac{(n-K)(n-2T)}{n(n-1)} \quad (A.5)$$

To guarantee $\mathbb{E}|\boldsymbol{\beta}\Delta\hat{\boldsymbol{\beta}}| \leq K\epsilon$, by solving (A.5) $\leq K\epsilon$, a passive design requires a minimal sample size of

$$T \geq \frac{p}{2} \geq \frac{n}{2}\left(1 - \frac{n-1}{n-K}\epsilon\right). \quad (A.6)$$

$\square$

**Corollary A.2.** *Using noiseless region-sensing observations, a passive design in 1D with $T \leq \frac{n}{2}$ region measurements achieves the optimal average-case Delta-risk, if and only if it can separate the search space into $2m$ disjoint subsets using intersections, unions, and complements of the measurement regions. The following example is adapted from Gray code:*

| $t$ | $\mathbf{x}^\top$ |
|---|---|
| 1 | 0 0 0 0 1 1 1 1 |
| 2 | 0 0 1 1 1 1 0 0 |
| 3 | 0 1 1 0 0 0 0 0 |
| 4 | 0 0 0 0 0 1 1 0 |
| $\cdots$ | $\cdots$ *(the pattern cycles)* |

(A.7)

*Proof.* To minimize (A.5), it is sufficient to find passive designs where $p = 2T$, given that the region aggregate measurements are noiseless. The expected risk of (A.5) turns out to be independent of the sizes of each elementary subset $C_j$ (which one can verify with a minimal example where $n = 4$, $K = 2$, and $p = 2$), which suggests that all designs that yield $p = 2T$ have the same average-case Delta-risk with noiseless region aggregate measurements. $\square$

Notice that the Gray-code design may not be optimal when the measurements are noisy. For this reason, we also included point sensing in Table 1 in our main paper, which yields the same order of sample complexity and performs better when the measurement noise is large.

# B Theoretical Properties for Active Sensing

The main goal of this section is to show that our main algorithm, *Region Sensing Index* (RSI), has the sample complexity guarantees show as Theorem 3 in the main paper. The main paper includes a proof sketch with three major steps. We show their details in 3 respective subsections.

## B.1 Basic Properties of Information Gain (IG)

Recall that the observation model is $y_t = \mathbf{x}_t^\top \boldsymbol{\beta} + \epsilon_t$, where $\boldsymbol{\beta} \in \mathcal{S}_\mu\binom{n}{k}$, $\boldsymbol{\beta} \sim \pi_t$, and $\epsilon_t \sim \mathcal{N}(0, \sigma_t^2)$. Omitting the time index $t$ in this subsection, the information gain (IG) to be maximized in every step is defined as

$$I(\boldsymbol{\beta}; y \mid \mathbf{x}, \pi) = H(y \mid \mathbf{x}, \pi) - \mathbb{E}[H(y \mid \mathbf{x}, \boldsymbol{\beta}) \mid \pi]$$
$$\Leftrightarrow \quad I(\gamma; y \mid \lambda, \mathbf{p}) = H(y \mid \lambda, \mathbf{p}) - H(\epsilon), \tag{B.1}$$

$$\text{where} \quad f(y \mid \lambda, \mathbf{p}) = \sum_{c=0}^{K} p_c \phi(y - c\lambda)$$

$$\lambda = \mu w_{\mathbf{x}}, \quad \gamma = \frac{\mathbf{x}^\top \boldsymbol{\beta}}{\lambda},$$

$$p_c = \Pr(\gamma = c) = \sum_{\boldsymbol{\beta}: \mathbf{x}^\top \boldsymbol{\beta} = c\lambda} \pi_t(\boldsymbol{\beta}). \tag{B.2}$$

There are two basic properties: Lemma B.1 that is both directly applied in Section 3.1 Accelerations and indirectly used in the later proof sketch; and Proposition B.2 that appears as Proposition 4 in the main paper.

**Basic Property 1**
**Lemma B.1** (Concavity and monotonicity). *$I(\gamma; y \mid \lambda, \mathbf{p})$ is concave in $\mathbf{p} \in \mathbb{R}_+^{K+1}$, which includes the convex simplex of $\Delta^K = \{\mathbf{p} \in [0,1]^{K+1} : \mathbf{p}^\top \mathbf{1} = 1\}$, if $0 < \lambda < \infty$ remains constant. On the other hand, $I(\gamma; y \mid \lambda, \mathbf{p})$ with fixed $\mathbf{p} \in \Delta^K$ is monotone-increasing as $\lambda$ increases.*

*Proof.* Concavity and monotonicity can be verified using derivatives. Notice the second term in (B.1) is constant. Here are the equations for the first term as well as its first and second order derivatives, omitting the dependency on $\mathbf{p}$ and $\lambda$ for simplicity:

$$H(y; \lambda, \mathbf{p}) = -\int f(y) \log f(y) \mathrm{d}y, \tag{B.3}$$

$$\partial H(y; \lambda, \mathbf{p}) = -\int \big(1 + \log f(y)\big) \partial f(y) \, \mathrm{d}y, \tag{B.4}$$

$$\partial^2 H(y; \lambda, \mathbf{p}) = -\int \left( \frac{\partial f(y) \partial f(y)^\top}{f(y)} + \big(1 + \log f(y)\big) \partial^2 f(y) \right) \mathrm{d}y \tag{B.5}$$

**Part 1.** To show concavity in $\mathbf{p}(\geq 0)$, let $\boldsymbol{\phi}_\lambda(y) = (\phi(y), \phi(y - \lambda), \ldots, \phi(y - K\lambda))^\top$ and write out the gradient and the Hessian of $H(y; \lambda, \mathbf{p})$ with respect of $\mathbf{p}$ as:

$$\frac{\partial H(y; \lambda, \mathbf{p})}{\partial \mathbf{p}^\top} = -\int_{-\infty}^{\infty} \big(1 + \log f(y)\big) \boldsymbol{\phi}_\lambda(y) \, \mathrm{d}y \tag{B.6}$$

$$\frac{\partial^2 H(y; \lambda, \mathbf{p})}{\partial \mathbf{p} \partial \mathbf{p}^\top} = -\int_{-\infty}^{\infty} \frac{1}{f(y)} \boldsymbol{\phi}_\lambda(y) \boldsymbol{\phi}_\lambda(y)^\top \mathrm{d}y \tag{B.7}$$

Notice $\boldsymbol{\phi}_\lambda(y) \boldsymbol{\phi}_\lambda(y)^\top$ is a PSD Gram matrix, which is preserved under integration. Further, the integral returns a PD matrix if the distribution is not degenerate ($\lambda > 0$ and $p_k > 0$ for at least two distinct $k$s)

**Part 2.1** For monotonicity in $\lambda(> 0)$, in the case when $K = 1$, the derivative with respect to $\lambda$ is

$$\frac{\partial H(y)}{\partial \lambda} = -\int \big(1 + \log f(y)\big) \cdot p_1 \phi(y - \lambda) \cdot (y - \lambda) \, \mathrm{d}y$$

$$= -p_1 \int \log f(y) \cdot \phi(y - \lambda) \cdot (y - \lambda) \, \mathrm{d}y$$

$$= -p_1 \int \log f(y + \lambda) \cdot \phi(y) \cdot (y) \, \mathrm{d}y, \tag{B.8}$$

where the first line removes constant integrals and the second shifts the variable. In order to show that (B.8) is nonnegative, pair up $y$ and $-y$ for $y > 0$ and notice that, by assuming $\lambda > 0$,

$$\phi(y + \lambda) \leq \phi(-y + \lambda) \Rightarrow f(y + \lambda) \leq f(-y + \lambda).$$

The bigger $\lambda$, the larger derivative it has.

**Part 2.2** In general when $K \geq 1$, we can write out the derivative as

$$\frac{\partial H(y; \lambda, \mathbf{p})}{\partial \lambda} = -\int_{-\infty}^{\infty} (1 + \log f(y)) \sum_{k=0}^{K} p_k \phi(y - k\lambda)(y - k\lambda) k \, dy$$

$$= -\sum_{k=1}^{K} \int_{-\infty}^{\infty} (1 + \log f(y)) \sum_{t=k}^{K} p_t \phi(y - t\lambda)(y - t\lambda) \, dy \quad \text{(B.9)}$$

Define $h_k(y) = \sum_{t=k}^{K} p_t \phi(y - t\lambda)$; we have

$$0 = -h_k(y) \log h_k(y) \Big|_{-\infty}^{\infty} = \int_{-\infty}^{\infty} (1 + \log h_k(y)) \sum_{t=k}^{K} p_t \phi(y - t\lambda)(y - t\lambda) \, dy \quad \text{(B.10)}$$

Consider each term of $k$ in (B.9) and add the corresponding terms from (B.10); using $\ell_k = \sum_{s=0}^{k-1} p_s \phi(y - s\lambda)$, we get

$$\frac{\partial H(y; \lambda, \mathbf{p})}{\partial \lambda} = -\sum_{k=1}^{K} \int_{-\infty}^{\infty} \log\left(1 + \frac{\ell_k(y)}{h_k(y)}\right) \sum_{t=k}^{K} \phi(y - t\lambda)(y - t\lambda) \, dy. \quad \text{(B.11)}$$

The only remaining task is to show that $r_k(y) = \frac{\ell_k(y)}{h_k(y)}$ is monotone decreasing with respect to $y$, which is sufficient to guarantee that (B.11) $\geq 0$, due to the odd symmetry of the remaining integrand parts around $y = t\lambda$. Take the derivative of $r_k(y)$ with respect to $y$:

$$r_k'(y) = \frac{\ell_k'(y) h_k(y) - \ell_k(y) h_k'(y)}{h_k^2(y)} = \frac{\sum_{s<k\leq t} p_s p_t \left(\phi_s'(y)\phi_t(y) - \phi_s(y)\phi_t'(y)\right)}{h_k^2(y)}$$

$$= \sum_{s<k\leq t} p_s p_t \frac{\phi_t^2(y)}{h_k^2(y)} \left(\frac{\phi_s(y)}{\phi_t(y)}\right)' = \sum_{s<k\leq t} p_s p_t \frac{\phi_t^2(y)}{h_k^2(y)} \left(\frac{\phi_s(y)}{\phi_t(y)}\right) \cdot (s\lambda - t\lambda) \leq 0, \quad \text{(B.12)}$$

where to simplify notations, we denote the composite function $\phi_s(y) = \phi(y - s\lambda)$. The inequality is strict if $\lambda > 0$ and $p_k \neq 0$ for at least two $k \in \{0, \ldots, K\}$. □

**Basic Property 2**
**Proposition B.2** (**Proposition 4** in the main document; a lower bound for the IG of a design). *The IG score of a region sensing design has lower bounds with respect to its design parameters $(\lambda, \mathbf{p})$, as*

$$I(\gamma; y \mid \lambda, \mathbf{p}) \geq 2q_c \bar{q}_c (2\Phi(\tfrac{\lambda}{2}) - 1)^2 \geq \tfrac{1}{12} \min\{q_c, \bar{q}_c\} \min\{\lambda^2, 3^2\}, \quad \forall 1 \leq c \leq K, \quad \text{(B.13)}$$

*where $q_c = P(\gamma \geq c) = \sum_{\kappa \geq c} p_\kappa$, $\bar{q}_c = 1 - q_c$, and $\Phi(u)$ is the standard normal cdf.*

*Proof of Proposition B.2.* **To show** (B.13), **first inequality:** Pick any $1 \leq c \leq K$; let $v = 1_{\gamma \geq c}$ and $\hat{v} = 1_{y > (c - 1/2)\lambda}$ be two binary truncations of the original variables, $\gamma$ and $y$, respectively. These truncations lose information:

$$I(\gamma; y \mid \mathbf{p}, \lambda) \geq I(v; \hat{v} \mid \mathbf{p}, \lambda) = \mathbb{E}^v K\big((\hat{v} \mid v) \| \hat{v}\big)$$

$$\geq 2 \sum_{v_0 \in \{0,1\}} P(v = v_0) \sup_{\hat{v}_0} \underbrace{\big|P(\hat{v} = \hat{v}_0 \mid v = v_0) - P(\hat{v} = \hat{v}_0)\big|^2}_{\triangleq \hat{\delta}(v_0, \hat{v}_0)}, \quad \text{(B.14)}$$

where $K(\cdot \| \cdot)$ is the Kullback–Leibler divergence and the second line comes from Pinsker's inequality.

Consider any realization of $v = v_0$ and choose $\hat{v}_0 = v_0$; using the rule of total probability and direct calculation,

$$\hat{\delta}(v_0, v_0) = \Big|P(\hat{v} = v_0 \mid v = v_0) - P(v = v_0) P(\hat{v} = v_0 \mid v = v_0) - P(v \neq v_0) P(\hat{v} = v_0 \mid v \neq v_0)\Big|$$

$$= P(v \neq v_0) \Big|P(\hat{v} = v_0 \mid v = v_0) - P(\hat{v} = v_0 \mid v \neq v_0)\Big|$$

$$\geq P(v \neq v_0) \left[\Phi\left(\frac{\lambda}{2}\right) - \left(1 - \Phi\left(\frac{\lambda}{2}\right)\right)\right] = P(v \neq v_0) \left(2\Phi\left(\frac{\lambda}{2}\right) - 1\right), \quad \text{(B.15)}$$

where $\Phi(\frac{\lambda}{2})$ is a lower bound on the probability of correct estimation, based on the worst-case draw of $\gamma$ such that $y$ cannot be more than $\frac{\lambda}{2}$ away from $\gamma$ in the direction that leads to estimation errors. Taking (B.15) to (B.14) yields the first part of the result.

**To show** (B.13), **second inequality:** So far we have shown an analytical lower bound for $I(\gamma; y \mid \lambda, \mathbf{p})$. To make the result even more interpretable, we can further numerically evaluate the Gaussian tail distribution, to find two constants, $C_1$ and $C_2$, such that

$$\Phi(x) - \frac{1}{2} = \int_0^x \phi(u)\,\mathrm{d}u = \int_0^x \frac{1}{\sqrt{2\pi}} e^{-\frac{u^2}{2}}\,\mathrm{d}u \geq C_1 \min\{x, C_2\}. \tag{B.16}$$

Since $\Phi(x)$ is monotone-increasing, we can fix $C_2$ to find the worst difference quotient, $\Phi(x)/x$, $\forall x \in (0, C_2]$. In fact, we can directly assign $C_1 \leq \Phi(C_2)/C_2$, because $\phi(u)$ is monotone-decreasing as $u$ increases. We choose $C_2 = \frac{3}{2}$ and $C_1 = \frac{1}{\sqrt{12}}$, which yields

$$\left(2\Phi\left(\frac{\lambda}{2}\right) - 1\right)^2 \geq \frac{1}{12}\min\{\lambda^2, 3^2\}. \tag{B.17}$$

□

**Proposition B.3** (An upper bound for the IG of a design). *When $K = 1$ and WLOG $p_1 \leq \frac{1}{2}$, the upper bound of IG derived from Jensen's inequality and max-entropy principle is $I(\gamma; y \mid \lambda, \mathbf{p}) \leq \frac{1}{2}p_1\lambda^2$, which is on the same order of (B.13) when $\lambda < O(1)$. In the $\lambda \gg 1$ case, the IG is naturally upper-bounded by a Bernoulli experiment with noiseless observation, $H(\mathcal{B}(p_1)) = -p_1 \log(p_1) - (1 - p_1)\log(1 - p_1) = \tilde{O}(p_1)$. Therefore, Proposition B.2 is a good approximation to the true IG in all scenarios. (See Figure 1 for an empirical visualization.) The general upper bound is not tight for general $k > 1$.*

*Proof.* The upper bound can be shown by Jensen's inequality and max-entropy principle. It is also tight when $k = 1$. Omitting $\mathbf{p}$ and $\lambda$,

$$I(\gamma; y) = I(\gamma; \gamma + \epsilon) = H(\gamma + \epsilon) + H(\gamma + \epsilon \mid \gamma) = H(\gamma + \epsilon) - H(\epsilon). \tag{B.18}$$

We only need to find the largest entropy for $H(\gamma+\epsilon)$ given $\mathbf{p}$ and $\lambda$. By Jensen's inequality, under the same mean and variance, a normal distribution has the largest entropy, where we have:

$$\mathbb{E}(\gamma + \epsilon) = \mathbb{E}(\gamma) + \mathbb{E}(\epsilon) = p_1\lambda, \quad \sigma^2_{\mathrm{mar}} = \mathrm{Var}(\gamma + \epsilon) = \mathrm{Var}(\gamma) + \mathrm{Var}(\epsilon) + 2\mathrm{Cov}(\gamma, \epsilon) = p_1\lambda^2 + 1. \tag{B.19}$$

We can then use a normal distribution with the above mean and variance as a upper bound to:

$$I(\gamma; y) = H(\gamma + \epsilon) - H(\epsilon) \leq \frac{1}{2}\log(2\pi e \sigma^2_{\mathrm{mar}}) - \frac{1}{2}\log(2\pi e) = \frac{1}{2}\log(1 + p_1\lambda^2) \leq \frac{1}{2}p_1\lambda^2 \tag{B.20}$$

□

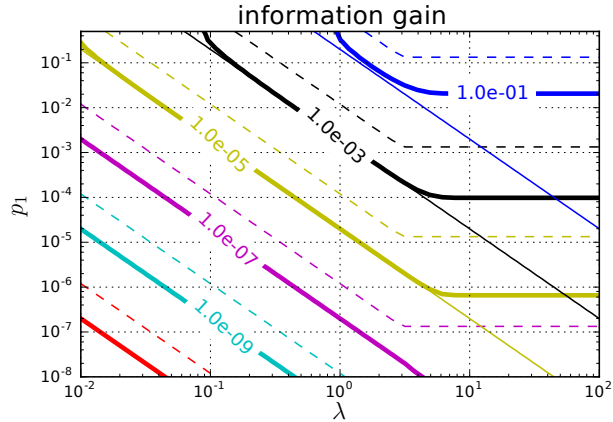

Figure 1: Level sets of IG $I(\gamma; y \mid \lambda, p_1)$ for different values of $p_1$ and $\lambda$, when $k = 1$. The thin lines below each true value indicate IG upper bounds (Proposition B.3) and the dashed lines are the phase-changing lower bound from Proposition B.2. The phase-changing bound is more useful because it produces insights about optimal region selection, usually at the point of phase-change, whereas the upper bound is non-informatively linear in the log-log plot.

## B.2 Minimum Information Gain of the Chosen Region in Each Iteration

This subsection aims to formalize the main observation in our main paper, which is that the information gain of all of the chosen measurements from RSI remain consistently large, before active search terminates with minimal Bayes risk. This observation implies a constant speed at which the model uncertainty can be reduced in expectation, leading to the upper bounds on sample complexity in Section B.3.

Recall that the Bayes risk is defined by $\bar{\epsilon}_t = \min_{|\hat{S}|=k} \frac{1}{k}\mathbb{E}[|\hat{S}\Delta S| \mid \pi_t]$, where $\Delta$ is the symmetric set difference operator. If we include the Bayes inference rule $\pi_t(\boldsymbol{\beta}) \propto \pi_0(\boldsymbol{\beta}) \prod_{\tau=1}^{t} p(y \mid \mathbf{x}_\tau, \boldsymbol{\beta})$, we can see that $\bar{\epsilon}_t$ is essentially a function of the collected data $D_t = \{(\mathbf{x}_\tau, y_\tau) : 1 \leq \tau \leq t\}$. The following lemma paraphrases Lemma 5 in the main document, with the time index $t$ omitted.

**Lemma B.4** (Minimum IG of the chosen regions). *WLOG, assume $n$ is a multiple of $2k$. At any step, given the data collection outcomes $D$ and the current Bayes risk $\bar{\epsilon}(D)$, we can always find a region $A$ of size at most $\frac{n}{k}$, such that $\lambda^2 \geq \frac{\mu^2}{a} = \frac{k\mu^2}{n}$ and $\frac{\bar{\epsilon}(D)}{2} \leq \mathbb{E}[\gamma_A \mid D] \leq 1 - \frac{\bar{\epsilon}(D)}{2}$ (we call it Condition E), which further yields*

$$I(\gamma; y \mid \lambda, \mathbf{p}) \geq I_{\bar{\epsilon}}^* = \frac{\bar{\epsilon}(D)}{25k} \min\{k\lambda^2, 3^2\} \geq \frac{\bar{\epsilon}(D)}{25k} \min\{\frac{k^2\mu^2}{n}, 3^2\}. \tag{B.21}$$

**Condition E** Lemma B.4 states the result in two steps: (a) the fact that the posterior model after collecting data $D$ still has large Bayes risk implies the existence of a very informative region that satisfies *Condition E* and (b) sensing on this region indeed yields nontrivial information, measured in terms of IG (B.21). We will split the proof into these two steps, accordingly.

**Lemma B.5** (A region that satisfies *Condition E*). *In 1d search with unit $\ell_2$-norm measurements, WLOG, assume $n$ is a multiple of $2k$. At any step, given the collected data $D$ and the current Bayes risk $\bar{\epsilon}(D)$:*

1. *There always is a region $B$ of size no larger than $\frac{n}{k}$, such that $\lambda_B^2 \geq \frac{\mu^2}{|B|} = \frac{k\mu^2}{n}$ and $\mathbb{E}[\gamma_B \mid D] \geq \frac{\bar{\epsilon}(D)}{2}$*

2. *There always is a subregion $A \subset B$ that satisfies* Condition E:

$$\lambda_A^2 \geq \frac{k\mu^2}{n} \quad \text{and} \quad \frac{\bar{\epsilon}(D)}{2} \leq \mathbb{E}[\gamma_A \mid D] \leq 1 - \frac{\bar{\epsilon}(D)}{2} \tag{B.22}$$

*Proof.* **Part 1.** Suppose the current minimizer of the posterior Bayes risk is $\hat{S} = \hat{S}(D) = \arg\max_{S'} \sum_{\hat{j} \in S'} \mathbb{E}[\beta_{\hat{j}} \mid D]$. Evenly split the domain into $K$ disjoint and contiguous regions and take their largest disjoint and contiguous subsets that do not intersect with $\hat{S}$. There are at most $G \leq 2K$ such sets; let them be $B_1, \ldots, B_G$. We use $\gamma(B_g) = \sum_{j \in B_g} \mathbf{1}_j^\top \boldsymbol{\beta}$ to denote the corresponding region latent variables in a region $B_g$. The region $B = \arg\max_{B_{g'}} \mathbb{E}[\gamma(B_{g'}) \mid D]$ yields

$$\mathbb{E}[\gamma(B) \mid D] \geq \frac{\sum_{g'=1}^{G} \mathbb{E}[\gamma(B_{g'}) \mid D]}{2K} = \frac{K - \mathbb{E}[\gamma(\hat{S}) \mid D]}{2K} = \frac{\bar{\epsilon}(D)}{2}, \tag{B.23}$$

due to the additivity, $\sum_g \gamma(B_g) + \gamma(\hat{S}) = \sum_{j \in [n]} \mathbf{1}_j^\top \boldsymbol{\beta} = K$.

**Part 2.** Let $A \subset B$ be the smallest contiguous subset such that $\mathbb{E}[\gamma(A) \mid D] \geq \frac{\bar{\epsilon}(D)}{2}$. Notice the maximum certainty of any point in $j \in A \subseteq \mathcal{X} \setminus \hat{S}$ is

$$\mathbb{E}[\beta_j \mid D] \leq \min_{\hat{j} \in \hat{S}}(1 - \mathbb{E}[\beta_{\hat{j}} \mid D]) \leq 1 - \bar{\epsilon}(D). \tag{B.24}$$

We then use the additivity of expectation to obtain

$$\mathbb{E}[\gamma(A) \mid D] \leq \mathbb{E}[\gamma(A \setminus \{j\}) \mid D] + \mathbb{E}[\beta_j \mid D] < \frac{\bar{\epsilon}(D)}{2} + (1 - \bar{\epsilon}(D)) = 1 - \frac{\bar{\epsilon}(D)}{2}, \quad \forall j \in A, \text{ i.e., } j \notin \hat{S}. \tag{B.25}$$

□

**IG of the Chosen Region** The following obtains Lemma B.6 with additive terms of $K^2$. It provides advantages over the straight-forward calculation in the main paper (which yields results with multiplicative factors of $K$).

**Lemma B.6** (Maximum IG when the outcome expectation is bounded). *For any design on $K$-sparse models, if there exists $0 < \bar{\epsilon} < 1$ and a design $(\mathbf{x}, A, \lambda, \gamma)$ such that $\frac{\bar{\epsilon}}{2} \leq \mathbb{E}\gamma \leq 1 - \frac{\bar{\epsilon}}{2}$, where $\gamma = \mathbf{x}^\top \boldsymbol{\beta}$ is latent variable of signal counts in the measurement region, then the information of the experiment is lower-bounded by*

$$I(\gamma; y \mid \mathbf{p}, \lambda) \geq \frac{\bar{\epsilon}}{25K} \min\{K\lambda^2, 3^2\} \tag{B.26}$$

*Proof.* We use the fact that IG is concave in **p** and we only check the vertices of the simplex of feasible probabilities to find its lower bound:

$$\begin{cases} p_k \geq 0, \quad k=1,\ldots,K, & \text{(Constraint } H_1,\ldots,H_K\text{)}; \\ \sum_{k=1}^{K} p_k \leq 1, & \text{(Constraint } H_0\text{)}; \\ \sum_{k=1}^{K} k p_k \geq \frac{\bar{\epsilon}}{2}, & \text{(Constraint } E_1\text{)}; \\ \sum_{k=1}^{K} k(1-p_k) \geq \frac{\bar{\epsilon}}{2}, & \text{(Constraint } E_2\text{)}, \end{cases}$$

(B.27)

where $p_0 = 1 - \sum_{k=1}^{K} p_k$ can be decided explicitly. All vertices of the simplex, including infeasible vertices, can be found by solving $K$ linear systems constructed from the $(K+3)$ linear constraints. Since $E_1$ and $E_2$ cannot be satisfied simultaneously for any $\bar{\epsilon} < 1$, we can enumerate all the remaining vertices and write out their respective nonzero values:

$$\begin{array}{lll} p_k = 1, & & \text{from } \cap_{k' \neq k} H_{k'}; \\ p_k + p_\ell = 1, & kp_k + \ell p_\ell = \frac{\bar{\epsilon}}{2}, & \text{from } \cap_{k' \neq k,\ell} H_{k'} \cap E_1; \\ p_k + p_\ell = 1, & kp_k + \ell p_\ell = 1 - \frac{\bar{\epsilon}}{2}, & \text{from } \cap_{k' \neq k,\ell} H_{k'} \cap E_2. \end{array}$$

(B.28)

The first row is infeasible when $\bar{\epsilon} > 0$. We then bound the IG for the other rows. Without loss of generality, assume $\ell < k$. Then, all feasible cases require $\ell = 0$ and yield $\min\{p_k, p_\ell\} \geq \frac{\bar{\epsilon}}{2k}$. Using Proposition 4,

$$I(v; u \mid \mathbf{p}, \lambda) \geq \frac{\bar{\epsilon}}{25K} \min\{K^2 \lambda^2, 3^2\} \geq \frac{\bar{\epsilon}}{25K} \min\{K\lambda^2, 3^2\}. \tag{B.29}$$

$\square$

*Proof of Lemma B.4.* The design from Lemma B.5 satisfies both *Condition E* and $\lambda \geq \frac{K\mu^2}{n}$, where we can then apply Lemma B.6 to obtain the conclusion. $\square$

## B.3 The Proof of Theorem 3

Lemma B.4 implies that the entropy in the posterior distribution, $H(\boldsymbol{\beta} \mid \pi_t) = -\sum_{\boldsymbol{\beta}} \pi_t(\boldsymbol{\beta}) \log \pi_t(\boldsymbol{\beta})$, decreases at least by $I_\epsilon^*$ with every measurement in expectation, starting with $H(\boldsymbol{\beta} \mid \pi_0) \leq k \log n$. Since the posterior entropy cannot be negative, RSI must terminates in finite times in expectation.

**Theorem B.7** (**Theorem 3** in the main document; sample complexity of RSI). *In active search of $k$ sparse signals with strength $\mu$ in 1d physical space of size $n(\geq 2k)$, given any $\epsilon > 0$ as tolerance of posterior Bayes risk, RSI using region sensing has bounded expected number of actual measurements before stopping,*

$$\bar{T}_\epsilon = \mathbb{E}[\min\{\mathcal{T} : \bar{\epsilon}(D_\mathcal{T}) \leq \epsilon\}] \leq 50\left(\frac{n}{\mu^2} + \frac{k^2}{9}\right) \log_2\left(\frac{2}{\epsilon}\right) \log\left(\frac{n}{\epsilon}\right) = \tilde{O}\left(\frac{n}{\mu^2} + k^2\right), \tag{B.30}$$

*where the expectation is taken over the prior distribution and sensing outcomes.*

**The Simple Approach**

**Definition B.8** (Stopping time). Define $T_\epsilon = \min_\mathcal{T}\{\bar{\epsilon}(D_\mathcal{T}) \leq \epsilon\}$ to be a random stopping time for an experiment to first yield less than $\epsilon$ posterior risk, $\bar{\epsilon}(D_\tau) = \frac{1}{K} \mathbb{E}[S\Delta\hat{S} \mid D_\tau] \leq \epsilon$. $T_\epsilon = T_\epsilon(\tau)$ can be determined given $\tau$.

**Lemma B.9** (Simple Expectations on the Number of Measurements for Small Errors). *Given any $\epsilon_1 > 0$, $t_0 \geq 0$, and the first $t_0$ data collection outcomes $D_{t_0}$, the expected number of additional measurements before the RSI stops with posterior risk less than $\epsilon_1$ is bounded in terms of $H_0 = H(\boldsymbol{\beta} \mid \pi_0)$ and $I_{\epsilon_1}$ defined in Lemma B.4, as*

$$\mathbb{E}(T_{\epsilon_1} - T_{\epsilon_0} \mid D_{t_0}) \leq \frac{H_0}{I_{\epsilon_1}} \leq \frac{25 H_0}{\epsilon} \max\left\{\frac{n}{k\mu^2}, \frac{k}{9}\right\}. \tag{B.31}$$

**Remark B.10.** *By taking $t_0 = 0$ and $H_0 \leq k \log n$, Lemma B.9 implies*

$$\bar{T}_\epsilon \leq \frac{25 \log(n)}{\epsilon_1} \max\left\{\frac{n}{\mu^2}, \frac{k^2}{9}\right\}. \tag{B.32}$$

*Proof of Lemma B.9.* Let $t = t_0 + s$ for any $s \geq 0$ and $D_t$ be the random variable for the data collection outcomes until step $t$. According to Lemma B.4,

$$(T_{\epsilon_1} \mid D_t) > t \quad \Rightarrow \quad H(\boldsymbol{\beta} \mid D_t) - \mathbb{E}^y\big[H(\boldsymbol{\beta} \mid D_t \cup \{\mathbf{x}, y\}) \mid D_t, \mathbf{x}_{t+1}\big] \geq I_{\epsilon_1} \tag{B.33}$$

$$\Rightarrow \quad H(\boldsymbol{\beta} \mid D_t) \geq I_{\epsilon_1} + \mathbb{E}^y\big[H(\boldsymbol{\beta} \mid D_t \cup \{\mathbf{x}, y\}) \mid D_t, \mathbf{x}_{t+1}\big] \tag{B.34}$$

Taking expectation over
$$\{D_t : (T_{\epsilon_1} \mid D_t) > t, D_{t_0}\} = \{D_t : \bar{\epsilon}(D_{t'}) > \epsilon_1, \forall t' \leq t, D_{t_0}\} \tag{B.35}$$
yields
$$\mathbb{E}[H(\boldsymbol{\beta} \mid D_t) \mid T_{\epsilon_1} > t, D_{t_0}] \geq I_{\epsilon_1} + \mathbb{E}[H(\boldsymbol{\beta} \mid D_{t+1}) \mid T_{\epsilon_1} > t, D_{t_0}], \tag{B.36}$$
where the expectation is taken over $(D_t \mid D_{t_0}, T_{\epsilon_1} > t)$ and $(D_{t+1} \mid D_{t_0}, T_{\epsilon_1} > t)$, respectively.

Next, we hope to apply Lemma B.4 at step $(t+1)$, but we have to make sure that the condition still holds, which is not directly implied by (B.36). To guarantee the conditions, we divide $D_{t+1}$ into two cases and use the nonnegativity of entropy to relax the second case,
$$\begin{aligned}
\mathbb{E}[H_{t+1}(\boldsymbol{\beta}) \mid T_{\epsilon_1} > t, D_{t_0}] &= P(T_{\epsilon_1} > t+1 \mid T_{\epsilon_1} > t, D_{t_0})\mathbb{E}[H_{t+1}(\boldsymbol{\beta}) \mid T_{\epsilon_1} > t+1, D_{t_0}] \\
&+ P(T_{\epsilon_1} = t+1 \mid T_{\epsilon_1} > t, D_{t_0})\mathbb{E}[H_{t+1}(\boldsymbol{\beta}) \mid T_{\epsilon_1} = t+1, D_{t_0}] \\
&\geq P(T_{\epsilon_1} > t+1 \mid T_{\epsilon_1} > t, D_{t_0})\mathbb{E}[H_{t+1}(\boldsymbol{\beta}) \mid T_{\epsilon_1} > t+1, D_{t_0}].
\end{aligned} \tag{B.37}$$

We can then iterate beginning with $t = t_0$ as
$$\begin{aligned}
\mathbb{E}[H_{t_0}(\boldsymbol{\beta}) \mid D_{t_0}] &\geq P(T_{\epsilon_1} > t_0 \mid D_{t_0})\mathbb{E}[H_{t_0}(\boldsymbol{\beta}) \mid T_{\epsilon_1} > t_0, D_{t_0}] \\
&\geq P(T_{\epsilon_1} > t_0 \mid D_{t_0})\bigg(I_{\epsilon_1} + P(T_{\epsilon_1} > t_0 + 1 \mid T_{\epsilon_1} > t_0, D_{t_0})\mathbb{E}[H_{t_0+1}(\boldsymbol{\beta}) \mid T_{\epsilon_1} > t_0 + 1, D_{t_0}]\bigg) \\
&= P(T_{\epsilon_1} > t_0 \mid D_{t_0})I_{\epsilon_1} + P(T_{\epsilon_1} > t_0 + 1 \mid D_{t_0})\mathbb{E}[H_{t_0+1}(\boldsymbol{\beta}) \mid T_{\epsilon_1} > t_0 + 1, D_{t_0}] \\
&\geq P(T_{\epsilon_1} > t_0 \mid D_{t_0})I_{\epsilon_1} + P(T_{\epsilon_1} > t_0 + 1 \mid D_{t_0})\bigg(I_{\epsilon_1}+ \\
&\quad + P(T_{\epsilon_1} > t_0 + 2 \mid T_{\epsilon_1} > t_0 + 1, D_{t_0})\mathbb{E}[H_{t_0+2}(\boldsymbol{\beta}) \mid T_{\epsilon_1} > t_0 + 2, D_{t_0}]\bigg) \\
&\geq P(T_{\epsilon_1} > t_0 \mid D_{t_0})I_{\epsilon_1} + P(T_{\epsilon_1} > t_0 + 1 \mid D_{t_0})I_{\epsilon_1} \\
&\quad + P(T_{\epsilon_1} > t_0 + 2 \mid D_{t_0})\mathbb{E}[H_{t_0+2}(\boldsymbol{\beta}) \mid T_{\epsilon_1} > t_0 + 2, D_{t_0}] \\
&\geq \ldots \\
&\geq I_{\epsilon_1}\sum_{s=0}^{\infty} P(T_{\epsilon_1} > t_0 + s \mid D_{t_0}) = I_{\epsilon_1}\mathbb{E}(T_{\epsilon_1} - T_{\epsilon_0} \mid D_{t_0}),
\end{aligned} \tag{B.38}$$
which leads to the conclusion given $\mathbb{E}[H_{t_0}(\boldsymbol{\beta}) \mid D_{t_0}] = H(\boldsymbol{\beta} \mid D_{t_0}) = H_0$. □

**The Complex Approach**

**Lemma B.11** (Max entropy given Bayes error). *For a K-sparse model, $\boldsymbol{\beta} \in \mathcal{S}\binom{n}{K}$, given $\bar{\epsilon} \geq \frac{1}{K}\sum_{j \in \hat{S}} P(\beta_j = 0) = \frac{1}{K}\mathbb{E}|S\Delta\hat{S}|$, the posterior entropy is at most*
$$H(\boldsymbol{\beta}) \leq KH(\mathcal{B}(\bar{\epsilon})) + K\bar{\epsilon}\log n, \tag{B.39}$$
$$\leq \frac{K}{2^r}(2r\log 2 + \log n), \quad \text{if } \bar{\epsilon} \leq \frac{1}{2^r}, \forall r = 0, 1, 2, \ldots \tag{B.40}$$
*where $H(\mathcal{B}(\bar{\epsilon})) = -\bar{\epsilon}\log\bar{\epsilon} - (1-\bar{\epsilon})\log(1-\bar{\epsilon})$ is denoted as the entropy of a Bernoulli experiment with $\bar{\epsilon}$ success rate.*

*Proof.* **Part 1.** Let $S = \{S_1, \ldots, S_K\}$ be the set of supports of the random variable $\boldsymbol{\beta}$ that is modeled by the posterior distribution given the history data that leads to the current state. We can compute the expectation as
$$\sum_{k=0}^{K} kP(|S\Delta\hat{S}| = k) = \mathbb{E}|\hat{S}\Delta S| \leq K\bar{\epsilon}. \tag{B.41}$$

Define $p_k = p_k(\hat{S}) = P(|S \Delta \hat{S}| = k)$; the total entropy can be bounded:

$$H(\boldsymbol{\beta}) = -\sum_{k=0}^{K} \sum_{S:|S \Delta \hat{S}|=k} \pi(\boldsymbol{\beta}_S) \log \pi(\boldsymbol{\beta}_S) \tag{B.42}$$

$$\leq -\sum_{k=0}^{K} p_k \log \left( \frac{p_k}{\binom{K}{K-k}\binom{n-K}{k}} \right) \tag{B.43}$$

$$\leq -\sum_{k=0}^{K} p_k \log p_k + \sum_{k=0}^{K} p_k \log \binom{K}{k} + \sum_{k=0}^{K} k p_k \log n \tag{B.44}$$

$$= -\sum_{k=0}^{K} p_k \log p_k + \sum_{k=0}^{K} p_k \log \binom{K}{k} + K \bar{\epsilon} \log n \tag{B.45}$$

where (B.42) separate the joint probabilities into $(K+1)$ groups according to their values of $|S \Delta \hat{S}|$. Inside every group, (B.43) realizes a uniform distribution, which maximizes the entropy given any value of group marginal probability, $p_k$. We then relax the number of combination by $\log \binom{x}{K-k} \leq (K-k) \log x$, which yields (B.44). From there, we use the condition, reformulated as (B.41), to obtain (B.45).

The next step uses the principle of maximum entropy to realize the optimizer for (B.45), when the moments are bounded by (B.41). The Lagrangian of the constrained optimization is

$$L(\mathbf{p}; c, \rho) = -\sum_{k=0}^{K} p_k \log p_k + \sum_{k=0}^{K} p_k \log \binom{K}{k} + c \left( \sum_{k=0}^{K} p_k - 1 \right) + \rho \left( \sum_{k=0}^{K} k p_k - K \bar{\epsilon} \right). \tag{B.46}$$

Setting the derivatives to zero yields

$$0 = \frac{\partial L}{\partial p_k} = -\log p_k + \log \binom{K}{k} + 1 + c + k\rho \quad \Rightarrow \quad p_k \propto \binom{K}{k} (e^\rho)^k, \tag{B.47}$$

which implies that $p_k$ is the probability of $k$ outcomes in a binomial distribution with $K$ rounds and an iid outcome probability of $p = \frac{1}{1+e^{-\rho}}$ in each round. Since the expectation of the total outcome is $K\bar{\epsilon}$, we have $p = \bar{\epsilon}$. Given the max-entropy binomial distribution and let $(X_1, \ldots, X_K)$ to be the outcome of each round; the entropy of their sum is upper bounded by the sum of their marginal entropies, which is $K$ times the entropy of $H(\bar{\epsilon})$. So, we proved (B.39).

**Part 2.** To move forward to (B.40), we need an interim result when $\bar{\epsilon} \leq \frac{1}{2}$:

$$H(\bar{\epsilon}) \leq -2\bar{\epsilon} \log \bar{\epsilon} \quad \Rightarrow \quad H(\boldsymbol{\beta}) \leq -2K\bar{\epsilon} \log \bar{\epsilon} + K\bar{\epsilon} \log N, \tag{B.48}$$

To show the interim result, let $\ell(\bar{\epsilon}) = -\bar{\epsilon} \log \bar{\epsilon} + (1 - \bar{\epsilon}) \log(1 - \bar{\epsilon})$; its derivatives are $\ell'(\bar{\epsilon}) = -\log \bar{\epsilon} - \log(1 - \bar{\epsilon}) - 2$ and $\ell''(\bar{\epsilon}) = -\frac{1}{\bar{\epsilon}} + \frac{1}{1-\bar{\epsilon}}$. The concavity of $\ell(\bar{\epsilon})$ in $0 \leq \bar{\epsilon} \leq \frac{1}{2}$ where $\ell''(\bar{\epsilon}) \leq 0$ and $\ell(0) = \ell(\frac{1}{2}) = 0$ yield $\ell(\bar{\epsilon}) \geq 0$, i.e., $H(\bar{\epsilon}) \leq -2\bar{\epsilon} \log \bar{\epsilon}, \forall 0 \leq \bar{\epsilon} \leq \frac{1}{2}$.

Finally, (B.40) trivially holds when $r = 0$. Otherwise, substitute $\bar{\epsilon} \leq 2^{-r}$ with $r \geq 1$ in (B.48) yields the final conclusion. $\square$

**Proof of the final theorem.** Let $\epsilon_r = 2^{-r}$ and $T_{\epsilon_r} = \min_{\mathcal{T}} \{\bar{\epsilon}(D_{\mathcal{T}}) \leq \epsilon_r\}$, for $r = 0, 1, \ldots, \lceil \log_2(\frac{1}{\epsilon}) \rceil$. From Lemma B.9, we have

$$\mathbb{E}(T_{\epsilon_{r+1}} - T_{\epsilon_r} \mid D_t, T_{\epsilon_r} \leq t) \leq \frac{H(\boldsymbol{\beta} \mid D_t)}{I^*_{\epsilon_{r+1}}}. \tag{B.49}$$

We can use Lemma B.4 with $\epsilon_{r+1} = 2^{-r-1}$ to show

$$I^*_{\epsilon_{r+1}} \geq \frac{\epsilon_{r+1}}{25k} \min\left\{ \frac{k^2 \mu^2}{n}, 9 \right\} \geq \frac{1}{50k 2^r} \min\left\{ \frac{k^2 \mu^2}{n}, 9 \right\} \tag{B.50}$$

and Lemma B.11 with $\bar{\epsilon}(D_t) \leq \epsilon_r = 2^{-r}$ to bound

$$H(\boldsymbol{\beta} \mid D_t) \leq \frac{k}{2^r}(2r \log 2 + \log n). \tag{B.51}$$

Put both bounds to (B.49) to get

$$\mathbb{E}(T_{\epsilon_{r+1}} - T_{\epsilon_r} \mid D_t, T_{\epsilon_r} \leq t) \leq 50 \max\left\{\frac{n}{\mu^2}, \frac{k^2}{9}\right\}(2r \log 2 + \log n). \tag{B.52}$$

Notice the right side is independent of $D_t$ and $t$, using linearity of expectations,

$$\mathbb{E}(T_{\epsilon_{r+1}} - T_{\epsilon_r}) \leq 50 \max\left\{\frac{n}{\mu^2}, \frac{k^2}{9}\right\}(2r \log 2 + \log n), \tag{B.53}$$

which further implies, using $R = \lceil \log_2 \frac{1}{\epsilon} \rceil < 1 + \log_2 \frac{1}{\epsilon}$,

$$\begin{aligned}
\mathbb{E}T_\epsilon &\leq \sum_{r=0}^{R-1} \mathbb{E}(T_{\epsilon_{r+1}} - T_{\epsilon_r}) \\
&\leq 50 \max\left\{\frac{n}{\mu^2}, \frac{k^2}{9}\right\} \sum_{r=0}^{R-1} (2r \log 2 + \log n) \\
&\leq 50 \max\left\{\frac{n}{\mu^2}, \frac{k^2}{9}\right\} R((R-1) \log 2 + \log n) \\
&\leq 50 \max\left\{\frac{n}{\mu^2}, \frac{k^2}{9}\right\} \log_2 \frac{2}{\epsilon} \log \frac{n}{\epsilon}
\end{aligned} \tag{B.54}$$

□

## C  Real-world experiments

For the purpose of more accurate modeling of the actual measurement powers at each region level, we use statistics from the real data to model $\mu w(a)$ and $\sigma(a)$ as functions of the region size $a$. The resulting SNR, $\lambda(a) = \frac{\mu w(a)}{\sigma(a)}$ is shown in Table 1. The optimal region size to begin under the uniform prior distribution is $32 \times 32$.

Figure 2 show examples of blue objects that are detected from the NAIP dataset.

Table 1: NAIP Blue Object Signal to Noise Ratio

| $\sqrt{a}$ | 1 | 2 | 4 | 8 | 16 | 32 | 64 | 128 |
|---|---|---|---|---|---|---|---|---|
| SNR | 8.5 | 8.4 | 7.7 | 5.6 | 2.7 | 0.9 | 0.3 | 0 |

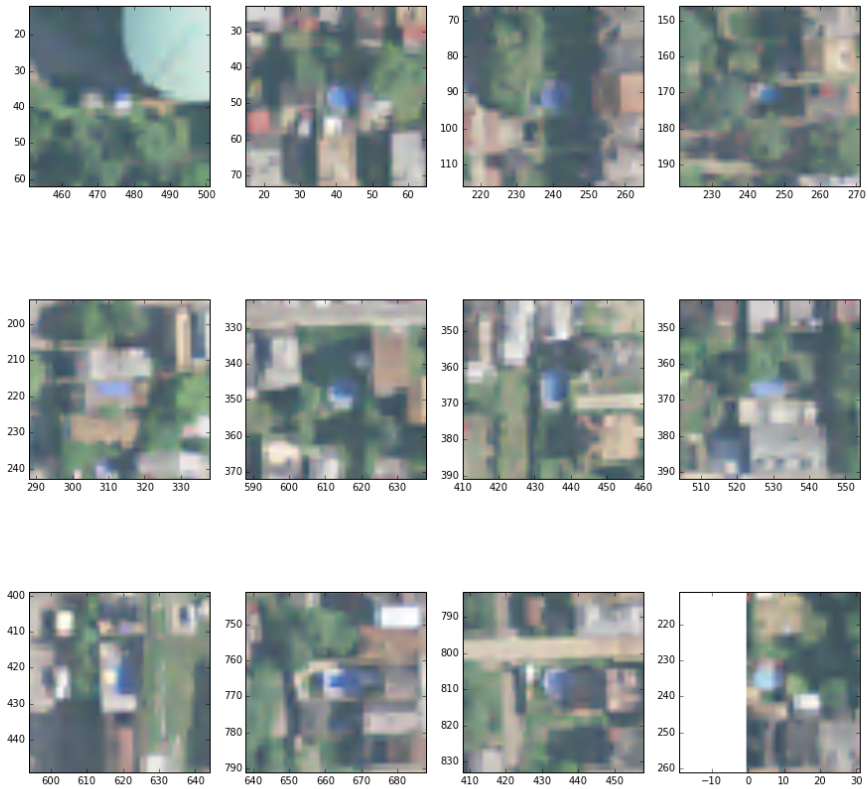

Figure 2: Example of positive discoveries in NAIP satellite images.